\documentclass[11pt]{article}

\usepackage[preprint]{acl}

\usepackage{times}
\usepackage{latexsym}
\usepackage{titlesec}
\usepackage[T1]{fontenc}

\usepackage[utf8]{inputenc}

\usepackage{microtype}

\usepackage{inconsolata}

\usepackage{graphicx}

\usepackage[table,dvipsnames]{xcolor}

\usepackage[utf8]{inputenc}
\usepackage[T1]{fontenc}
\usepackage{times}
\usepackage{microtype}
\usepackage{subcaption}
\usepackage{amsmath, amssymb, amsfonts, amsthm}
\usepackage{latexsym}

\usepackage{algorithm}
\usepackage{algpseudocode}

\usepackage{graphicx}
\usepackage{wrapfig}

\usepackage{booktabs}
\usepackage{siunitx}
\usepackage{multirow}
\usepackage{tabularx}
\usepackage{makecell}
\usepackage{threeparttable}

\usepackage{enumitem}

\definecolor{mygray}{gray}{.9}

\definecolor{darkblue}{RGB}{0, 0, 209}

\title{Cache Mechanism for Agent RAG Systems}

\newcommand*\samethanks[1][\value{footnote}]{\footnotemark[#1]}

\author{
  Shuhang Lin\textsuperscript{1}\thanks{Equal contribution.} \quad
  Zhencan Peng\textsuperscript{1}\samethanks \quad
  Lingyao Li\textsuperscript{2} \quad
  Xiao Lin\textsuperscript{3} \quad
  Xi Zhu\textsuperscript{1} \quad
  Yongfeng Zhang\textsuperscript{1}\\
  \textsuperscript{1}Rutgers University \quad
  \textsuperscript{2}University of South Florida \quad
  \textsuperscript{3}University of Illinois Urbana--Champaign \\
  \texttt{\{shuhang.lin, zhencan.peng, xi.zhu, yongfeng.zhang\}@rutgers.edu} \\
  \texttt{lingyaol@usf.edu, xiaol13@illinois.edu}
}

\begin{document}
\maketitle
\begin{abstract}
Recent advances in Large Language Model (LLM)-based agents have been propelled by Retrieval-Augmented Generation (RAG), which grants the models access to vast external knowledge bases. Despite RAG's success in improving agent performance, agent-level cache management, particularly constructing, maintaining, and updating a compact, relevant corpus dynamically tailored to each agent's need, remains underexplored. Therefore, we introduce \textbf{ARC (Agent RAG Cache Mechanism)}, a novel, annotation-free caching framework that dynamically manages small, high-value corpora for each agent.  By synthesizing historical query distribution patterns with the intrinsic geometry of cached items in the embedding space, ARC automatically maintains a high-relevance cache. With comprehensive experiments on three retrieval datasets, our experimental results demonstrate that ARC reduces storage requirements to \textbf{$0.015\%$} of the original corpus while offering up to \textbf{$79.8\%$} has-answer rate  and reducing average retrieval latency by \textbf{$80\%$}. Our results demonstrate that ARC can drastically enhance efficiency and effectiveness in RAG-powered LLM agents. 
\end{abstract}

\begin{figure*}[!t]  
  \centering
  \includegraphics[width=\textwidth,keepaspectratio]{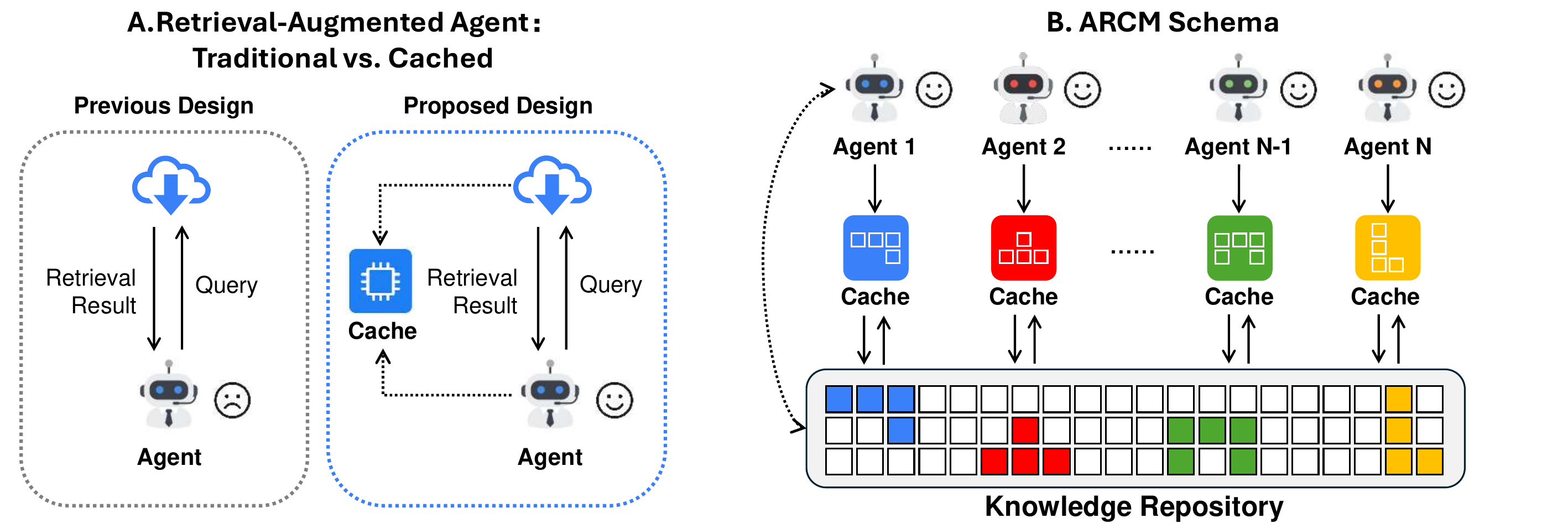}
  \caption{The motivations of our research: (a) The comparison between traditional and cached retrieval-augmented agent. (b) Our proposed ARCM schema.}
  \label{fig:example}
\end{figure*}

\section{Introduction}
LLM agents have demonstrated significant potential in various domains, which exhibit remarkable capabilities in performing complex reasoning tasks \cite{wang2024sibyl,jin2024exploring, hua2024disentangling}, executing knowledge-intensive operations \cite{jiang2024kgagent}, and implementing customized automated workflows \cite{li2024autoflow,xu2025mem, mei2025litecua}. Researchers have employed multiple techniques to enhance these agents' ability, including RAG, finetuning, and prompt engineering. Among these approaches, RAG empowers agents with the capacity to access extensive external knowledge repositories, facilitating more informed response generation and consequently becoming widely implemented across agent architectures \cite{chujie2024honestllm,huang2025trustworthiness,guo2025deepsieve}.

The enhancement of LLM agent capabilities has led to significant growth in knowledge repositories, with extensive-scale knowledge bases now prevalent in advanced systems. With the accumulation of diverse and multifarious documents, several hundred GB or TB-level corpora are common among enterprise and government entities. These huge knowledge repositories not only enable LLM agents to address a more comprehensive range of inquiries, but also bring the following challenges: (1) high storage costs, and (2) deployment difficulties on edge and mobile devices.

Firstly, massive knowledge repositories drive up hardware and maintenance expenses—storing terabytes of embeddings and documents demands high-performance disks or distributed storage clusters—and complicate system engineering due to the need for sophisticated indexing, sharding, and cache invalidation logic \cite{mei2502omnirouter,mei2025eccos}. Furthermore, deploying RAG at the edge or on mobile devices is particularly challenging: such platforms have limited local storage and CPU/GPU resources, and often operate under unreliable or high-latency network conditions. Every off-device retrieval incurs tens to hundreds of milliseconds in round-trip delay, degrading user experience and potentially causing service unavailability during network outages. These constraints motivate the introduction of an efficient, high-speed caching layer that stores only the most semantically central passages. By dramatically reducing both remote calls and on-device compute, a well-managed cache cuts latency, lowers bandwidth consumption, and simplifies the overall architecture—eliminating the need to bundle or host the full knowledge base on every device.


While the constraints outlined above motivate efficient caching solutions, recent approaches have fallen short. Numerous studies have attempted to introduce caching mechanisms to accelerate LLM systems, yet the majority of these approaches primarily focus on key-value caches, which incur significant storage costs and consequently limit their applicability in the very resource-constrained environments they aim to serve. These conventional caching strategies operate independently of semantic relationships between queries and documents, thus failing to fully leverage the semantic-aware retrieval techniques that make modern RAG systems effective in the first place.

Our insight brings guidance to cache algorithm design. We propose the \textbf{Agent RAG Cache Mechanism (ARC)}, which introduces two innovative components: (1) a rank-distance based frequency that weights query relevance by both occurrence frequency and semantic similarity to top-ranked results, and (2) a centrality quantification through hubness, where items with higher neighbor connectivity intuitively contain more general knowledge relevant to the agent's domain. We combine these two factors to create a priority score for cached items that governs cache maintenance. On a large-scale corpus comprising millions of embedding–passage pairs, ARC achieves a \textbf{79.8\% cache has-answer rate} while using just \textbf{0.015\%} of the original storage.

In summary, our main contributions are as follows:
\begin{itemize}[topsep=0pt,itemsep=1pt,parsep=0pt,partopsep=0pt]
\item \textbf{Problem.} To the best of our knowledge, this work presents the first dedicated caching mechanism designed for retrieval-augmented LLM agents, and establishes a formulation of this problem for analyzing cache efficiency and performance.

\item \textbf{Methodology.}
  We propose \textbf{ARC}, an agent RAG caching algorithm that leverages query-based dynamics and the structural properties of the item representation space, which can drastically reduce storage requirements while preserving retrieval effectiveness.

  \item \textbf{Evaluation.} We evaluate ARC on three datasets, with the 6.4-million-document Wikipedia as a realistic, noisy external corpus. Compared to prior methods, ARC attains a cache size of \textbf{0.015\%} of the full index, a \textbf{79.8\%} cache has-answer rate, and an \textbf{80\%} average reduction in retrieval latency.

\end{itemize}


\section{Related Work}
\textbf{RAG for agent.} RAG enables agents to efficiently access external knowledge. Early studies embedded one‐off retrieval calls directly into prompts for single‐step decision‐making or multi‐hop question answering, allowing LLMs to reference external documents during generation and thereby improving the accuracy of individual responses or action instructions \cite{shi-etal-2024-generate, lee2024planrag, huang2025breaking, guo2025deepsieve}. Subsequently, frameworks such as PlanRAG \cite{lee2024planrag} proposed a phased pipeline, which decomposes complex tasks into subgoals \cite{guo2025reagannodeasagentreasoninggraphagentic}, retrieves pertinent information for each subtask, and synthesizes the results in a unified generation step; this reduces redundant queries and enhances the traceability and efficiency of multi‐step decision processes \cite{lee2024planrag,huang2024datagen}. In multi‐hop reasoning contexts, the Generate‐then‐Ground \cite{shi-etal-2024-generate} paradigm was introduced: the model first generates intermediate hypotheses, which are then grounded via document‐level verification by the retrieval module, improving the robustness of chain‐of‐thought reasoning \cite{shi-etal-2024-generate}. More recently, methods such as RAP \cite{kagaya2024rap} and RAT \cite{wang2024rat} have incorporated mechanisms for contextual memory and dynamic retrieval triggering, enabling agents in multimodal or long‐horizon interactions to invoke external knowledge adaptively based on internal confidence metrics or new sensory inputs, thus providing flexible, on‐demand information supplementation. Cutting‐edge frameworks like RAG‐Gym employ fine‐grained process supervision to jointly optimize retrieval and reasoning workflows \cite{xiong2025raggym}.

\textbf{Efficient RAG. }
Efficient Retrieval-Augmented Generation aims to reduce resource consumption while improving end-to-end performance. Research in this field can be organized into three major areas: (1) Retrieval algorithm optimizations, including sparse indexing such as IVF/PQ \cite{johnson2017billion} and low-dimensional vector representations with approximate nearest neighbor search \cite{quinn2025accelerating}; (2) Pipeline optimizations, such as multi-stage cascaded retrieval, which progressively refines candidate sets to focus expensive ranking on high-recall subsets \cite{bai2024pistis}; and (3) Hardware-aware compression techniques, such as 4-bit embedding quantization for RAG \cite{jeong2025_4bit}, alongside comprehensive evaluations of compression and dimensionality-reduction trade-offs \cite{huerga2025optimization}.

\textbf{Semantic Encoding.} The field of text embedding has progressively advanced from static representations such as Word2Vec \citep{mikolov2013distributed} to contextualized models like ELMo \citep{peters2018deep} and BERT \citep{devlin2019bert}, which generate dynamic word vectors conditioned on surrounding context. Building upon these frameworks, Siamese-network and contrastive-learning approaches such as SBERT \citep{reimers2019sentence} and SimCSE \citep{gao2021simcse} produce high-quality sentence and paragraph embeddings, while retrieval-augmented methods like DPR \citep{karpukhin2020dense} and ColBERT \citep{khattab2020colbert} integrate explicit document retrieval with dense vector representations. More recent work including LM-Steer \citep{han2023word} has explored mechanisms for task- or style-guided embedding control. Concurrently, intrinsic challenges of high-dimensional spaces have been widely recognized, including the curse of dimensionality \citep{bellman1957dynamic}, hubness \citep{radovanovic2010hubs}, and anisotropy \citep{ethayarajh2019towards}. These challenges have motivated a range of mitigation strategies, including local scaling, isotropy regularization, and manifold-alignment techniques.

Despite advances in retrieval systems, two critical gaps remain: the unexploited hierarchical nature of embeddings within topic clusters, and the neglect of embedding-space geometry in existing neural-retriever caches. We address these limitations by (1) formalizing embedding stratification to identify semantically central passages, and (2) designing a cache mechanism that leverages both spatial geometry and agent-specific semantics. Our approach combines hubness-based centrality with dynamic rank-distance metrics to enable efficient RAG retrieval in LLM agents.

\section{Problem Formulation}

In this section, we cast the agent’s RAG pipeline as a constrained optimization problem, and further develop a detailed formulation that factorizes the pipeline into three sequential stages: (1) query generation, (2) retrieval search, and (3) cache storage.

\textbf{Query generation.} 
Unlike general-purpose conversational LLMs, agents are typically associated with a predefined domain specialization that specifies their intended application context \citep{russell2016artificial}. This specialization induces domain-specific response preferences. For instance, a medical agent tends to provide precise answers to clinical queries \cite{ely2005answering}. Once users become aware of such preferences, their interactions often exhibit implicit biases: users are more likely to ask medical agents health-related questions rather than queries from unrelated domains \cite{ebell1999information}, such as astrophysics. This latent query bias implies that an agent's query distribution is shaped by its operational context. Formally, this relationship can be expressed as $q \sim \mathcal{P}(q|\Theta)$, where $q$ is the user query, $\Theta$ refers to the agent's domain specialization, and $\mathcal{P}$ denotes the probability distribution of queries conditioned on the agent's domain specialization.

\textbf{Retrieval Search.} In the retrieval phase, a RAG-enhanced agent leverages the user-provided query to identify highly relevant documents from an external corpus. Given a query $q$ and an external corpus $U$, the set of documents retrieved by the agent can be formally defined as:
$$
\operatorname{Ret}(q, U) = \{ x_i : x_i \in \operatorname{arg\ TopK}_{x_i \in U} \left( \operatorname{sim}(q, x_i) \right) \},
$$
where $x_i$ denotes a candidate retrievable item (e.g., document passage) from the corpus $U$, and $\operatorname{sim}(q, x)$ is the agent’s internal similarity function that scores the relevance between the query $q$ and each item $x$.

\textbf{Cache storage.} As the agent engages in frequent interactions with users, the cache utilization of a RAG-enhanced system tends to grow continuously due to the need to retrieve new documents over time. However, retaining all retrieved content is impractical, as it would require prohibitively large cache capacity. In real-world deployments, it is therefore critical to design RAG agents that operate efficiently under limited cache budgets. Let $w(x_i)$ denote the cache space occupied by the $i$-th retrieved item. The constraint of operating within a fixed cache budget naturally introduces a capacity constraint into the optimization problem:
\begin{equation*}
    \sum_{x_i\in C_t} w(x_i) \;\le\; W_{max}, \forall t \geq 0
\end{equation*}
where $W_{max}$ is the max capacity of the cache. Let $C_t$ denote the set of items stored in the cache at time $t$.

\textbf{Cache Metrics.}
To quantify query-conditioned cache efficiency, we define the cumulative miss count for query \( q_t \) and the empirical has-answer rate over \( T \) queries as  
$$
M_t = \sum_{j=1}^m \mathbb{I}[x_{t,j} \notin C_{t,j-1}]
$$

$$
\text{Has-AnswerRate}_T = 1 - \frac{1}{mT} \sum_{t=1}^T M_t
$$
Here, $\mathbb{I}(\cdot)$ serve as the indicator function.
\( x_{t,j} \) denotes the \( j \)-th item in query \( q_t \), and \( C_{t,j-1} \) is the cache state before accessing it. The has-answer rate reflects the average fraction of items served from the cache, offering a proxy objective for optimizing learned caching policies based on retrieval success.

\subsection{Agent RAG Cache: From Query to Cache}

Let $q_t$ denote the $t$-th query sampled from a distribution $\mathcal{P}(q\,|\,\Theta)$. Prior to processing $q_t$, the cache is in state $C_{t-1} \subseteq U$, constrained by a total capacity $\sum_{x \in C_{t-1}} w(x) \le W_{\max}$, where $w(x)$ represents the memory footprint of item $x$.

The retrieval module returns a ranked list of candidates:
\[
R_t = \operatorname{Ret}(q_t, U) = \left(x_{t,1}, x_{t,2}, \dots, x_{t,m}\right),
\]
with each $x_{t,j}$ selected from the top-$K$ corpus items under a similarity function:  
\[
x_{t,j} \in \operatorname{TopK}_{x \in U} \left( \operatorname{sim}(q_t, x) \right).
\]

Upon accessing candidate $x = x_{t,j}$, the augmented candidate pool becomes $D = C_{t,j-1} \cup \{x\}$. The cache state is then updated via selection:$
C_{t,j} = \arg\max_{S \subseteq D} \sum_{y \in S} \, p(y) $ subject to, $\sum_{y \in S} w(y) \le W_{\max},
$
where $p(y)$ denotes the utility score guiding cache prioritization. If the candidate is redundant or infeasible—i.e., $x \in C_{t,j-1}$ or $w(x) > W_{\max}$—no update is applied: $C_{t,j} = C_{t,j-1}$. After all $m$ candidates are considered, the final cache state is set as $C_t = C_{t,m}$. $S$ ranges over all subsets of the augmented candidate set $D$.

\subsection{Optimization Objective}

At inference step $n$, our objective is to optimize the cache policy—specifically, the priority function $p(\cdot)$ and update rule—so as to maximize the expected has-answer rate across a future horizon of $H$ queries:
\begin{equation*}
\begin{split}
\max_p\; &\mathbb{E}_{q_{n+1:n+H} \sim \mathcal{P}(q|\Theta)}
   \bigl[\text{Has-AnswerRate}_{n+1:n+H}\bigr] \\
&= \max_p\; \mathbb{E} \left[ 1 - \frac{1}{mH}
   \sum_{t=n+1}^{n+H} M_t \right].
\end{split}
\end{equation*}
This formulation defines our core goal: to design a cache strategy that maximizes future query performance by learning from historical retrieval patterns, subject to fixed capacity constraints.

\section{Agent RAG Cache Mechanism}

In this section, we introduce ARC, a cache algorithm designed for RAG-powered agents. ARC maintains the cache by deriving two complementary factors into a unified \emph{priority} value for each cache item. Our ARC algorithm leverages both query-based dynamics and the structural properties of the item representation space: while the distance–rank frequency score (DRF) quantifies the dynamic demand for passages demonstrated through historical queries, the hubness score identifies passages within the cache that are inherently more likely to be retrieved due to their inherent spatial structure. By synthesizing these complementary signals, we establish a balanced item prioritization mechanism that adapts to agent query distribution patterns while ensuring cache priorities remain grounded in the geometric properties of the passage embedding space.

\subsection{Distance--Rank Frequency Score}
While a significant number of caching algorithms are designed with a heavy dependence on item access frequency, these approaches usually lack consideration of retrieval system characteristics. In these frequency-based methods, all retrievals of a passage typically contribute equally to its cache importance score. However, when a passage appears as a top-ranked result, it generally provides more value than another passage appearing in the tenth position for the identical query. In other words, higher-ranked appearances indicate stronger semantic alignment with queries and should be weighted more heavily in caching decisions. By implementing differential weighting based on retrieval rank, our caching mechanism incorporates rank position as a key signal in our priority formulation, creating a more retrieval-aware caching strategy.

Rank-based weighting improves cache prioritization by emphasizing higher-ranked retrievals, which captures the relative order of results. However, we still need to reflect actual semantic closeness between queries and passages. For example, a tenth-rank hit with high cosine distance should score lower than another tenth-rank result that is semantically closer. To address this, we also incorporate the absolute distance between query and passage embeddings into our scoring mechanism. By combining both rank for ordinal importance and distance for semantic proximity, our priority formula delivers a more nuanced evaluation of each passage's cache value.

\paragraph{Distance--Rank Frequency (DRF) Score.}
Let $\mathcal{Q}$ denote the set of cached queries and for each query $q\in\mathcal{Q}$ let $\operatorname{Ret}(q)$ be the ranked retrieval set (top-$k$) returned from the cache or external database. Define the DRF score of item $p$ as
$$\mathrm{DRF}(p) 
= \sum_{q\in\mathcal{Q}\colon p\in\mathrm{\operatorname{Ret}(q)}} 
\frac{1}{\mathrm{rank}(q,p) \;\cdot\; \mathrm{dist}(q,p)^{\alpha}}
$$
where $\mathrm{rank}(q,p)$ denotes the 1-based position of $p$ in the retrieval results for query $q$, $\operatorname{dist}(q, p)$ represents the embedding distance between $q$ and $p$, and $\alpha>0$ is a tunable parameter controlling distance sensitivity.

In summary, DRF factor accumulates a passage's retrieval history, weighted by both its rank position and embedding distance for each query. It therefore directly captures how frequently and meaningfully the passage responds to the current query distribution, reflecting real-time demand and relevance in the agent system.

\subsection{Hubness Score}

In this subsection, we introduce the hubness score, computed on the nearest neighbor search index of the cache candidate set. This metric highlights passages in the dense cores of the embedding space, specifically those that frequently appear in other embeddings' neighbor lists, without relying on any query history. As a result, it provides a truly query-agnostic approach for identifying which cache entries are most semantically central and broadly useful.

\paragraph{Space-Aware Approach.}
Hubness refers to the statistical phenomenon in high-dimensional spaces where certain points called "hubs" appear disproportionately often in others' k-nearest-neighbor lists \cite{tomasev2013role}. Prior studies have examined hubness from algorithmic and statistical perspectives but have paid little attention to its semantic implications. 

Existing works \cite{bogolin2022cross,wang2023balance} rrevealthat passages with high hubness scores consistently exhibit higher retrieval rates across diverse query sets, reflecting a space-invariant measure of semantic centrality. This insight motivates the first component of our cache mechanism. However, while hubness effectively captures the global importance of items, our formulation must also consider local retrieval dynamics. We therefore introduce both rank and absolute distance components into our prioritization formula. The rank component addresses the varying importance of different positions in retrieval results, while absolute distance proves crucial in practice; for instance, when query-document similarity is exceptionally low, such results should receive correspondingly reduced weighting regardless of rank.

Furthermore, due to sharing the same mathematical formulation in top-K selection, hubness serves as a natural bridge between retrieval behavior and embedding space characteristics, enabling us to simulate retrieval dynamics without making assumptions about the query distribution.

\textbf{Hubness Score.} 
Hubness score of $x$ counts its occurrences in other points' $k$-nearest-neighbor lists. Which in math can be written as $$
h_k(x_i) = \sum_{\substack{j=1 \\ j \neq i}}^{n} \mathbb{I}(x_i \in \mathcal{N}_k(x_j))
$$

$\mathcal{N}_k(x_j)$ denotes the k-nearest neighbors of $x_j$, $\mathbb{I}(\cdot)$ serve as the indicator function. In the ARC algorithm, we compute hubness over the cache’s own candidate set.

\paragraph{Combined Priority.}
Our priority calculation integrates two key metrics: DRF and the hubness score introduced in Section 3. With an additional penalty term for memory utilization. Using $w(p)$ to denote the memory footprint of item $p$, we define:
\begin{equation*}
\begin{split}
\mathrm{Priority}(p)
&= \frac{1}{\log\!\bigl(w(p)+1\bigr)}
 \Bigl[\beta\,\log\!\bigl(h_k(p)+1\bigr) \\
&\qquad\quad + (1-\beta)\,\operatorname{DRF}(p)\Bigr].
\end{split}
\end{equation*}
where $\beta\in[0,1]$ balances centrality versus query frequency. The $\log(w(p) + 1)$ term in the denominator penalizes items with large memory footprints and encourages the cache to store more items. Items with lower $\mathrm{Priority}(p)$ are deemed less valuable and evicted first to make room for new insertions.

Our comprehensive formula accounts for global importance via hubness, relative positional relationships via rank, and absolute semantic proximity via distance. By integrating these multiple dimensions of retrieval relevance, our approach offers superior performance compared to frequency-based formulations alone. This provides a more nuanced mechanism for determining cache priority that aligns with the semantic characteristics of embedding-based retrieval systems and remains robust across varying query distributions.
\subsection{Cache Maintenance}
Algorithm~\ref{alg:arcache} shows the maintenance procedure of ARC managing a cache $C$ with maximum capacity $W_{max}$ and escalating the query to $\mathcal{D}$. When a new query $q$ arrives, the algorithm first retrieves the top-$k$ candidates $Ret(q) \gets \mathrm{topK}(q, C)$ from the cache. The relevance of these results is evaluated by computing the mean embedding distance $\mathrm{avg}_{p\in Ret(q)}(\mathrm{dist}(q,p))$ across all retrieved items. Only if the average distance exceeds $\tau$, indicating low relevance, ARC escalates retrieval to the full external corpus $\mathcal{D}$ and updates the retrieved items $Ret(q) \gets \mathrm{topK}(q, \mathcal{D})$. 

Subsequently, ARC updates the distance-rank metrics of each item $p \in Ret(q)$. To simplify the notation, we defined $\mathrm{DR}(p,q) = 1/{\mathrm{rank}(q,p) \cdot \mathrm{dist}(q,p)^{\alpha}}.$ Existing cache entries accumulate their DRF score via $\mathrm{DRF}(p) \mathrel{+}= \mathrm{DR}(p,q),$ while new items initialize $\mathrm{DRF}(p) = \mathrm{DR}(p,q)$ before being inserted into $C$. Finally, cache $C$ iteratively evicts the lowest-priority item $\arg\min_{x\in C}\mathrm{Priority}(x)$ until the total cache weight satisfies $\sum_{x\in C}w(x) \leq W_{\max}$.

\begin{algorithm}[t]
\caption{Agent RAG Cache Maintenance and Escalation}\label{alg:arcache}
\textbf{Input}: Query vector $q$, External corpus $\mathcal{D}$, Cache $C$ (with capacity $W_{\max}$), Top-$k$ retrieval size $k$, Drift tolerance $\delta$ \\
\textbf{Output}: Top-K Items $Ret(q)$
\begin{algorithmic}[1]
  \State $Ret(q) \gets Ret(q, C)$  \Comment{Return the topK result from the cache}
  \If{$\mathrm{avg}_{p\in Ret(q)} (\mathrm{dist}(q,p)) > \tau$}
    \State $Ret(q) \gets Ret(q, U)$   \Comment{Return the topK result from the external corpus U}
  \EndIf

  \For{each $p \in Ret(q)$}    \Comment{Update the DSF Score of inserted/existed items in cache}
    \If{$p \in C$}
      \State $\mathrm{DRF}(p)  \mathrel{+}= DR(p,q)$  
    \Else
      \State $\mathrm{DRF}(p)=DR(p,q)$       
      \State $C \gets C \cup \{p\}$           
    \EndIf
  \EndFor

  \While{$\sum_{x\in C} w(x) > W_{\max}$}     \Comment{Evict if cache exceeds capacity}
    \State $C \;\gets\; C \setminus \bigl\{\arg\min_{x\in C}\mathrm{Priority}(x)\bigr\}$
  \EndWhile

  \State \Return $Ret(q)$
\end{algorithmic}
\end{algorithm}

\section{Empirical Evaluation}

\begin{figure*}[t]
    \centering
    \begin{subfigure}[b]{0.48\textwidth}
        \centering
        \includegraphics[width=\linewidth]{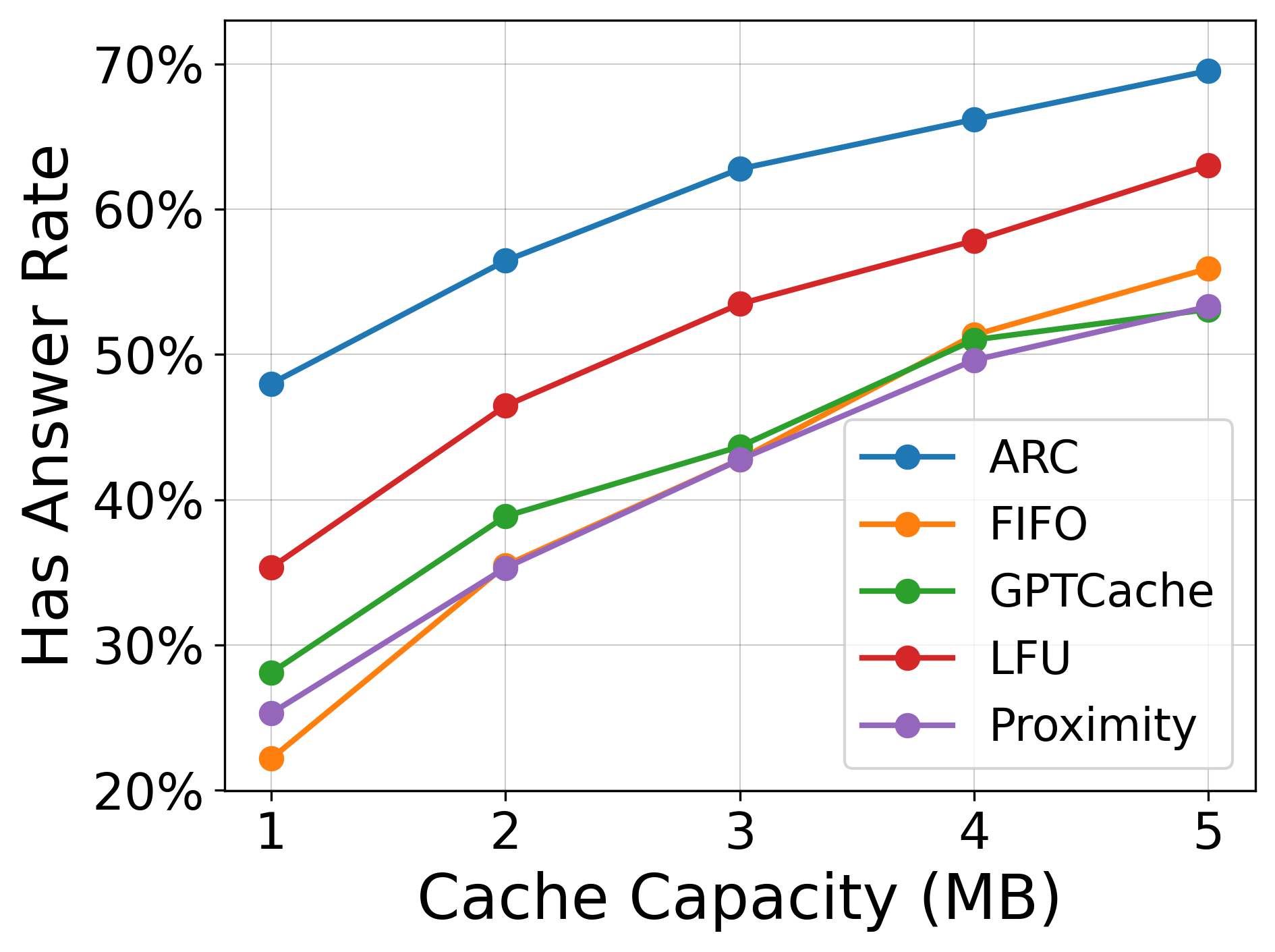}
        \subcaption{Has-answer Rate at Varying Cache Capacity}
        \label{fig:ablation_cache_capacity}
    \end{subfigure}
    \hfill
    \begin{subfigure}[b]{0.48\textwidth}
        \centering
        \includegraphics[width=\linewidth]{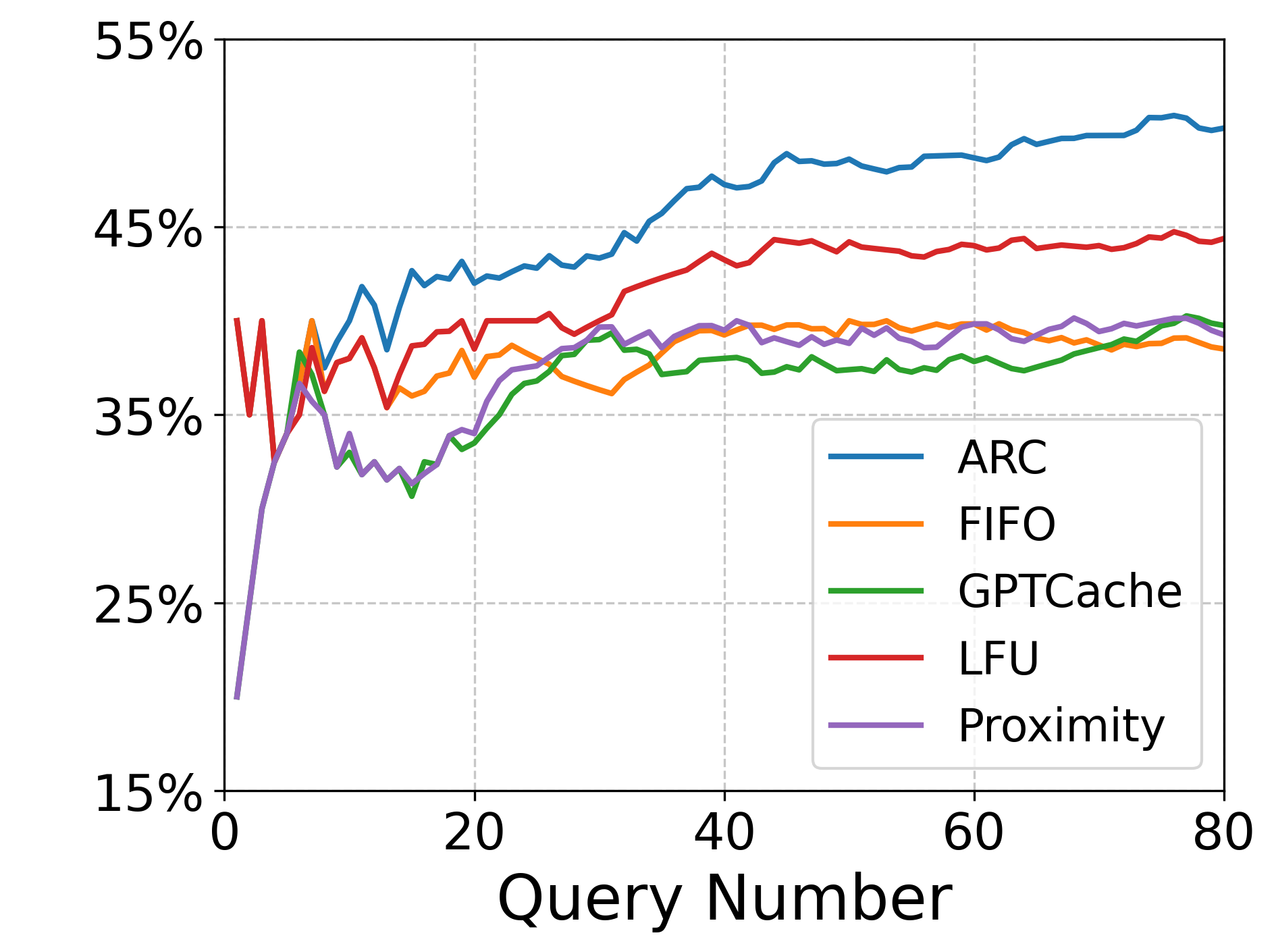}
        \subcaption{Cache Has-answer Rate Across Query Stream}
        \label{fig:stream_has_answer_rate_in_mmlu}
    \end{subfigure}
    \caption{Cache performance analysis: (a) Effects of varying cache capacity; (b) Continuous improvement of has-answer rate with streaming queries.}
    \label{fig:cache_analysis}
\end{figure*}

\textbf{Datasets:}
\textbf{Large-scare Retrieval Corpus.}
To simulate a real-world, large-scale RAG deployment, we build our primary index from the English subset of the 2023 Wikipedia dump\footnote{\url{https://huggingface.co/datasets/wikimedia/wikipedia}} which has over 6.4 million documents. We segment each document in passage level, where each chunk have one or more passages with no more than 2048 characters. The total chunk amount is over 14 milion. This full-scale document repository introduces substantial distractor noise—forcing the retriever to discriminate finely between relevant and irrelevant content. We also inject into this index all passages containing the ground-truth answers for our evaluation sets to evaluate performance. 

\textbf{Datasets: Evaluation Datasets.}
We assess ARC on three diverse question-answering datasets including SQuAD, MMLU and Adversarial QA.

We evaluate our method on three datasets: \textbf{SQuAD}~\cite{rajpurkar-etal-2016-squad}, \textbf{MMLU}~\cite{hendrycks2020measuring}, and \textbf{AdversarialQA}~\cite{jia-liang-2017-adversarial}. SQuAD contains 536 short Wikipedia articles yielding 107{,}785 QA pairs, where answers are contiguous spans within each context. MMLU spans 57 academic and professional subjects, with 15{,}908 test and 1{,}540 calibration questions. AdversarialQA is built on the SQuAD v1.1 dev set (10{,}570 examples) and introduces perturbations via AddSent and AddOneSent distractors.

\textbf{Baselines.} We compare our cache method with several standard policies, including LFU, FIFO, Proximity, and GPTCache. LFU and FIFO are the most classic baselines in cache-related methods: LFU evicts the least frequently used items, while FIFO removes the oldest entries in the cache. \textbf{Proximity}~\cite{bergman2025leveraging} maintains historical query–document pairs and returns previously retrieved passages from the most semantically similar past query when the similarity exceeds its threshold~$\tau$, evicting the oldest query–document pair. \textbf{GPTCache}~\cite{bang2023gptcache} reuses results from semantically similar queries as an eviction mechanism; however, its eviction strategy relies solely on embedding similarity, neglecting valuable retrieval-specific signals such as rank-weighted frequency or embedding-space centrality that could better capture the demand intensity and semantic centrality of cached items. Optimal parameter configurations for both Proximity and GPTCache were determined as $\tau = 0.2$ through systematic grid search. In ARC, we set $\alpha = 0.4$, and $\beta = 0.7$, $0.15$, and $0.2$ for the SQuAD, MMLU, and AdversarialQA datasets, respectively. Unless stated otherwise, all experiments use a cache capacity of 3.0~MB and a top-K retrieval size of $K=50$.

\textbf{Models.}
For the embedding model, given that we need to index all wiki content, utilizing closed-source models would impose significant computational costs. Therefore, following the comprehensive RAG evaluation\citep{Wang2024SearchingFB}, we adopted the open-source embedding models bge-small-en\cite{baai2023bgesmallen} and llm-embedder\cite{baai2023llmembedder} for our indexing architecture,  resulting in an embedding memory footprint of approximately 20 GB and 40 GB, respectively. We employed FAISS IndexFlatIP \cite{douze2024faiss} for our vector similarity search.

\textbf{Metrics.}
To evaluate the effectiveness of our cache system, we measure performance from two critical aspects: efficiency and latency reduction. We employ Has-Answer Rate, which measures the proportion of queries that can be successfully answered using cached content without accessing the full index, where a higher has-answer rate indicates better cache utilization and retrieval efficiency. Additionally, we use Average Memory Access Time (AMAT) \cite{hennessy2011computer}, which quantifies the average time required to retrieve information, accounting for both cache hits and misses. This metric effectively captures the overall latency reduction achieved by our caching mechanism compared to the baseline retrieval system.

\begin{table*}[t]
  \centering
  \caption{Has-Answer Rate$\uparrow$ (\%)}
  \label{tab:hit_rate_combined}
  \small 
  \begin{tabular}{@{}l ccc ccc@{}}
    \toprule
    \multirow{2}{*}{Method} 
      & \multicolumn{3}{c}{bge-small-en} 
      & \multicolumn{3}{c}{llm-embedder} \\
    \cmidrule(lr){2-4} \cmidrule(lr){5-7}
      & MMLU   & AdversarialQA & SQuAD  
      & MMLU    & AdversarialQA & SQuAD   \\
    \midrule
    LFU       & 53.26 & 66.11 & 69.46 & 42.09 & 57.57 & 59.37 \\
    FIFO      & 41.84 & 66.64 & \underline{77.04} & 34.51 & \underline{58.67} & \underline{71.49} \\
    GPTCache  & 43.65 & 46.49 & 41.23 & 40.08 & 41.54 & 40.13 \\
    Proximity & 46.41 & 54.25 & 59.79 & 38.41 & 50.73 & 58.84 \\ 
    ARC (w/o hubness) & \underline{62.37} & \underline{68.14} & 76.09 & \underline{51.92} & 58.48 & 69.55 \\
    ARC       & \textbf{62.63} & \textbf{71.18} & \textbf{79.80}
              & \textbf{52.46} & \textbf{62.78} & \textbf{71.94} \\
    \bottomrule
  \end{tabular}
\end{table*}

\begin{table*}[t]
  \centering
  \caption{AMAT$\downarrow$ (s)}
  \label{tab:amat_combined}
  \small
  \begin{tabular}{@{}l ccc ccc@{}}
    \toprule
    \multirow{2}{*}{Method}
      & \multicolumn{3}{c}{bge-small-en}
      & \multicolumn{3}{c}{llm-embedder} \\
    \cmidrule(lr){2-4} \cmidrule(lr){5-7}
      & MMLU & AdversarialQA & SQuAD
      & MMLU & AdversarialQA & SQuAD \\
    \midrule
    LFU       & 0.668 & 0.427 & 0.388 & 1.754 & 1.219 & 1.175 \\
    FIFO      & 0.814 & 0.429 & \underline{0.298} & 1.969 & \underline{1.182} & \underline{0.821} \\
    GPTCache  & 0.788 & 0.704 & 0.773 & 1.810 & 1.680 & 1.720 \\
    Proximity & 0.758 & 0.604 & 0.539 & 1.840 & 1.406 & 1.198 \\
    ARC (w/o hubness) & \underline{0.559} & \underline{0.411} & 0.307 & \underline{1.489} & 1.196 & 0.878 \\
    ARC       & \textbf{0.556} & \textbf{0.377} & \textbf{0.269}
              & \textbf{1.468} & \textbf{1.046} & \textbf{0.818} \\
    \bottomrule
  \end{tabular}
\end{table*}

\textbf{Baselines Comparison.}
To evaluate the performance of ARC, we conduct experiments comparing it against baselines including LFU, FIFO, Proximity, GPTCache, and ARC without the hubness component, across multiple retrieval benchmarks using two embedding models: bge-small-en and llm-embedder. We report the has-answer rate and AMAT in Table \ref{tab:hit_rate_combined} and Table \ref{tab:amat_combined}. \emph{Best results are \textbf{bolded}; second-best are \underline{underlined}.}

As shown in Table \ref{tab:amat_combined}, it is observed that GPTCache and Proximity perform worst among all methods. This is because their effectiveness relies on the assumption that incoming queries are highly similar to previously seen ones, which only holds under extensive historical data and large cache. In contrast, ARC highest has-answer rates and the lowest AMAT across all benchmarks. For example, under the bge-small-en embedding, even without the hubness score, ARC achieves a has-answer rate of over 62\% and an AMAT of 0.559s on MMLU, while LFU only achieves 53\% and 0.668s. With the hubness score, ARC further improves by over 3\% has-answer rate with hubness on Squad. Similarly, ARC outperforms all baselines in the llm-embedder setup, attaining ranging from 52.46\% to 71.94\%. The advantage of ARC comes from two combined score: DRF, which helps the cache identify valuable items based on their ranking position and semantic proximity to queries, and hubness score which naturally favors passages that hold central or influential positions within the embedding space. It is remarked that without caching, retrieving the entries with the highest semantic similarity on the Wikipedia index takes 1.313s per query on SQuAD, while ARC only needs 0.269s, reducing the query time by almost 80\%.

\textbf{Cache Capacity Ablation.} Figure \ref{fig:ablation_cache_capacity} illustrates cache has-answer rates varying cache capacity from 1MB to 5MB in Dataset MMLU with embedder bge-small-en. ARC consistently maintains the highest has-answer rates across various cache sizes, which significantly outperform baseline methods, particularly at smaller cache capacities. For example, ARC achieves 47.64\% has-answer rate at 1 MB of cache capacity on MMLU, surpassing LFU by 12.8\%.

\textbf{Streaming Query.} Figure~\ref{fig:stream_has_answer_rate_in_mmlu} shows the streaming cache has-answer rates over 80 sequential queries on MMLU. ARC quickly adapts to the incoming stream and consistently maintains the highest cumulative has-answer rate among all methods. While all baselines experience an initial warm-up phase, the has-answer rate of ARC continues to improve and stabilizes above 50\% after around 60 queries, significantly outperforming all alternatives such as LFU and Proximity.

\section{Conclusion}
 We introduce ARC, an annotation-free cache construction mechanism using stratified hub selection and adaptive escalation. On a 30 million-pair corpus, ARC achieves a 79.8\% has-answer rate while caching only 0.015\% of the data, outperforming baseline methods. By exploiting the hierarchical structure of embeddings, ARC provides a principled framework for data-efficient retrieval in bandwidth and latency-constrained RAG systems, opening avenues for research on geometric properties of representation spaces.

\section{Limitations} We focus our evaluation on the standard single-turn QA setting, which allows controlled measurement of caching behavior. While multi-turn dialogue is outside our current scope, ARC is compatible with session-level extensions; exploring dialogue-aware scoring and evaluation on conversational benchmarks is a natural next step.
We have not yet performed experiments in transfer or cross-domain settings. Extending ARC to evaluate transfer robustness will be a direction for future work.




\bibliography{References}



\end{document}